\documentclass[letterpaper]{article}

\usepackage{natbib,alifeconf}  %% The order is important
\usepackage{amsmath}

\usepackage{url,hyperref,cleveref}
\usepackage{booktabs}
\usepackage{caption}
\usepackage{listings}
\lstdefinestyle{mystyle}{
    basicstyle=\usefont{T1}{fvm}{m}{n}\fontsize{5}{5}\selectfont,
    basewidth=0.5em,
    breakindent=0pt,
    breakatwhitespace=true,   
    columns=fullflexible,
    breaklines=true,                 
    keepspaces=true,                 
}
\usepackage{dblfloatfix}

\usepackage{tabularray}
\SetTblrInner{
    rowsep = 0pt,  
    abovesep = 0pt,
    belowsep = 0pt
}
\UseTblrLibrary{siunitx}
\usepackage{siunitx}

\usepackage{inconsolata}
\newcommand{\smalltt}[1]{{\footnotesize\texttt{#1}}}
\newcommand{\ssmalltt}[1]{{\scriptsize\texttt{#1}}}
\newcommand\blfootnote[1]{%
  \begingroup
  \renewcommand\thefootnote{}\footnote{#1}%
  \addtocounter{footnote}{-1}%
  \endgroup
}

\usepackage[all=normal,wordspacing=tight]{savetrees}

\title{Emergent Culture in Minimal LLM Systems}

\author{
    Simon Jones$^{1}$ \and
    Sabine Hauert$^{1}$
    \mbox{}\\
    $^1$University of Bristol, UK \\
    simon2.jones@bristol.ac.uk
} % email of corresponding author

\begin{document}
\maketitle

\begin{abstract}
What happens when LLM agents operate with no context outside a turn, minimal prompting, and simple tools? Inspired by swarm engineering, we give collectives of three agents the ability to send messages and manipulate a shared actively decaying text store, introducing evolutionary pressure. The agents spontaneously cooperate, develop storage management strategies, and generate complex evolving cultural artifacts, with no top-down engineering. Using tools from dynamical systems analysis, we show that these behaviours exhibit structured long-range coherence beyond the entropy horizon of the decaying store, consistent with emergent culture in the Sperberian sense.
\end{abstract}

% Choose one of: Full Paper, Summaries, or Late Breaking Abstracts 
%Submission type: \textbf{Full Paper}\\

% If sharing code / data, anonymize your repository and paste the link here.
% Example of anonymizing sevice for github: https://anonymous.4open.science/
% delete this line if not needed
% Data/Code available at: \url{https://anonymous.4open.science/r/swarm_llm-402B}
Data/Code available at: \url{https://bitbucket.org/hauertlab/swarm_llm}
\blfootnote{\textcopyright  2026 Simon Jones and Sabine Hauert. Published under a Creative Commons Attribution 4.0 International (CC BY 4.0) license.}

\section{Introduction}
Most work on collaborating LLMs has focussed on more, more context, more memory, more prompt engineering, more infrastructure. But what if we take a step back and build a system stripped back to a minimum, giving stateless models access to some shared decaying storage and basic communications and a minimal prompt, and nothing else? Would we get emergent complexity and interesting behaviours? Or would systems of such agents collapse into repetition or stasis?

Remarkably, the answer is the former. Multiple runs with different model combinations reliably generate role allocation, coordination protocols, world-building, myth-making, and increasing complexity over time. By stripping the agents context back to the minimum, they are forced to keep external records, to fight entropy, to develop systems, to evolve culture. Although the individual outputs are often wildly different, there are deep similarities and some fundamental attractors emerge.

We draw on lessons from swarm robotics, and swarms in nature, where collectives of simple agents using minimal communication via the environment or message passing produce complex emergent system-level behaviours. The mode of communication via external decaying storage is a form of stigmergy; as termites build cathedrals, so stateless swarm-LLMs build coherent narratives, a signal of culture as defined by Sperber, where agents must continually read and recreate persistent meaning.

\section{Related work}
Much work on building multi-agent LLM systems has focussed on building external infrastructure and scaffolding; \cite{park2023generative} build a structured memory system, with retrieval, reflection, and planning, and agents have a pre-constructed persona, \cite{li2023camel} have communicating agents, each with a constructed role, \cite{hong2023metagpt} constrain agents with standard operating procedures and careful prompting, \cite{packer2023memgpt} construct an OS inspired memory hierarchy to manage limited context. In each case, the task-specific process is approached as a top-down design problem, messiness and unreliability to be managed and constrained. Work on using LLMs to generate large texts is generally heavily curated and focussed on hierarchical scaffolding \citep{bai2024longwriter,teleki2025survey,wang2025generating}. It is observed, but under-researched in the literature, that model performance degrades to repetition and information gain over multi-turn interactions decreases \citep{hong2025context,gorle2025quantifying}, generating more than a few thousand words of coherent text remains difficult. 
\cite{perez2026llmsplaytelephonegame} use iterated tasks to study LLM attractors, with single stateless agents successively modifying input texts.

The field of swarm robotics \citep{brambilla2013swarm} by contrast, relies on bottom-up emergence of complex large-scale behaviours from the noisy interactions of individual agents with each other and with the environment. Communication via the environment, stigmergy \citep{grasse1959reconstruction} is an important organising principle \citep{heylighen2016stigmergy}. We take inspiration from this; by stripping away almost all top-down scaffolding, and providing a noisy, entropic shared memory communication channel, we construct an environment where structure emerges naturally, and is sustained.

We know that far-from-equilibrium systems exhibit emergent complex structure \citep{prigogine1978time}, with large scale persistent organisation. We argue that placing agents in an environment with an entropic sink, the decaying shared memory, forces them to use `energy' in the form of the token flow to maintain structure. This persistent structure, lasting beyond the entropy horizon after which all older data is corrupt or deleted is a signal of culture in the Sperberian sense \citep{sperber1996explaining,claidiere2014darwinian} of continual reconstruction, agents must read, re-interpret, summarise, and write to reset the entropy clock. This is not cultural transmission \citep{dawkins1976selfish} because there is no perfect reproduction, agents reconstruct biassed by their cognitive attractors. Sperber's framework predicts that stable cultural artifacts are those that survive reconstruction through noisy channels and biassed minds. 

We show this using several levels of analysis; from low level dataflow analysis highlighting the differing emergent storage management strategies, the generation and persistence of novel vocabulary over long periods, to large scale semantic trajectories. Using semantic embeddings \citep{li2022brief,lin2017structured,giorgi2021declutr} of the generated text divided into sections, we use techniques from dynamical systems analysis; Recurrence Plots \citep{eckmann1987recurrence} and RQA \citep{zbilut1992embeddings,marwan2007recurrence}.

\section{Methods}
We use the models listed in Table \ref{tab:models}. All three models are capable of interleaved reasoning
and tool use, and all are frontier models, though Gemini has greater emphasis on long context (1M tokens) and speed, while Claude and Kimi have greater focus on reasoning.

We construct a minimal system such that stateless agents interact with each other and with a shared entropic key-value store, over a number of cycles. We use the LiteLLM Python API to provide an agnostic interface to provider endpoints. 

At the start of a cycle, each agent is invoked using the standard completions interface provided by LiteLLM, with a system prompt, a user prompt, and a tool schema. The tools available to the agents are shown in Table \ref{tab:tools}. All messages and storage data writes and appends are strings only, appends to a storage key result in a list of strings. Writing to storage and sending messages consumes a per-agent, per-cycle character budget, those tool calls return the characters consumed and the remaining budget, or a failure and the remaining budget if there is insufficient budget.
Each agent is given the same system prompt `You are a curious and creative agent. Explore and play!'. They are then given the user prompt `Cycle [n]. Continue your work.'. No previous context from earlier conversation turns is provided at all, the only persistent elements are the store and the messages in an agent's inbox.

\begin{figure}
    \centering
    \includegraphics[width=0.85\linewidth]{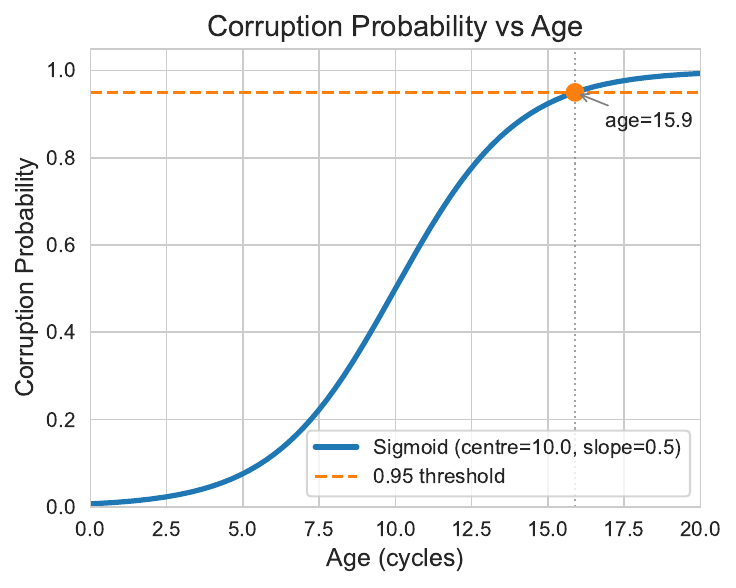}
    \caption{Sigmoid corruption probability. Alphanumeric characters within storage elements undergo probabilistic progressive corruption based on their age. Once $p>0.95$, the element is deleted.}
    \label{fig:sigmoid}
\end{figure}

Once the completions call has been sent to the model endpoints, the agents may make multiple tool calls in parallel and in sequence, interleaved with thinking and content output. All this happens asynchronously, though individual storage actions are atomic. After all agents have completed their turn, the shared memory  and agent message inboxes are subject to entropy. With a sigmoid probability, Figure \ref{fig:sigmoid}, based on the age of a storage element in cycles since it was written or sent, alphanumeric characters are corrupted, while whitespace and punctuation is preserved. Once the sigmoid reaches 95\%, the element is deleted. The sigmoid midpoint is 10 cycles and the slope is 0.5, so elements older than 3 cycles are likely to be corrupted, and around 16 cycles will be deleted. Appending to a key creates a new storage element, with its own age and does not affect the age of earlier elements. Finally, the cycle is complete and the next cycle starts.

We trial all two model combinations for ten cycles over ten runs to map the characteristics of the systems. These experiments, using a character budget of 20kchars showed that while Kimi and Gemini were broadly comparable in their token use and Kimi about twice as verbose as Gemini, Claude used far more tokens and generated far more text than Kimi and Gemini, 3.7x and 7.2x respectively. For the main experiment, we conduct runs with three agents over 100 cycles, with two model combinations, each repeated five times. The agent combinations for this were Kimi-Kimi-Gemini (kkg), and Claude-Kimi-Gemini (ckg). Model combination choice was driven by wanting to always have a long context window model and two reasoning models. The original character budget was reduced to a quarter, at 5kchars, more in line with the typical output level of Kimi and Gemini, in order to prevent excessive token use by Claude. The output and telemetry generated by these runs constitute the main data analysed.  
\begin{table}   
    \centering \caption{Specific models used}\label{tab:models}
    \begin{tblr}{stretch=0.9}
    \hline
         Company & Model & Nickname\\
         \hline
         Anthropic& \smalltt{claude-sonnet-4-5-20250929} & claude\\
         Moonshot& \smalltt{kimi-k2-thinking-turbo}  & kimi\\
         Google& \smalltt{gemini-3-flash-preview} & gemini\\
         \hline
    \end{tblr}
    \label{tab:placeholder}
\end{table}
\section{Results}
\subsection{Analysis methodology}
Analysing the outputs of these experiments is non-trivial, each run, even short ones, produces large amounts of text, structured in various ways. The primary output artifact is the event log, comprising a JSON file with every event that occurred, with timestamps and metadata, available in the repository.

To approach the analysis of the three-agent 100 cycle runs, we use a three layer approach. Firstly, we look at the actual dataflow between the agents and storage, noting common emergent communication patterns. Then we perform a textual analysis, looking at novel vocabulary persisting beyond the entropy horizon, and finally we use text embeddings and RQA analysis to quantitively position the agent output with respect to some examples of the human corpus, and to look at long-range semantic correlations.
\begin{table}[!t]
    \centering \caption{Tool schema. The system and user prompts, together with the text here is the entirety of the context that an agent is presented with, outside what they create and read. Return values from a tool call sent to agent are in JSON format}\label{tab:tools}
    \begin{tblr}{width=\linewidth,colspec={p{3.5cm} X[l]},stretch=0.8}
    \hline
    Tool & Description as seen by agent  \\
    \hline
    \textbf{\ssmalltt{get\_my\_id()}} & \textbf{\ssmalltt{Get your ID}} \\
    \textbf{\ssmalltt{list\_agents()}} & {\setlength{\baselineskip}{0.7\baselineskip}\textbf{\ssmalltt{Discover agents you can collaborate with}}}\\ 
    \textbf{\ssmalltt{list\_storage\_keys()}} & \textbf{\ssmalltt{List keys}} \\ 
    \textbf{\ssmalltt{read\_inbox(unread\_only=true)}} & \textbf{\ssmalltt{Read messages}} \\
    \textbf{\ssmalltt{send\_message(to, content)}} & \textbf{\ssmalltt{Send message}} \\
    \textbf{\ssmalltt{read\_storage(key)}} & \textbf{\ssmalltt{Read value}} \\
    \textbf{\ssmalltt{write\_storage(key, value)}} & {\setlength{\baselineskip}{0.7\baselineskip}\textbf{\ssmalltt{Write value, overwrites any existing value}}} \\
    \textbf{\ssmalltt{append\_storage(key, value)}}& \textbf{\ssmalltt{Append value}} \\
    \hline
    \end{tblr}
\end{table}
\subsection{Dataflow behaviour}
\begin{figure}[!t]
    \centering
    \includegraphics[width=0.9\linewidth]{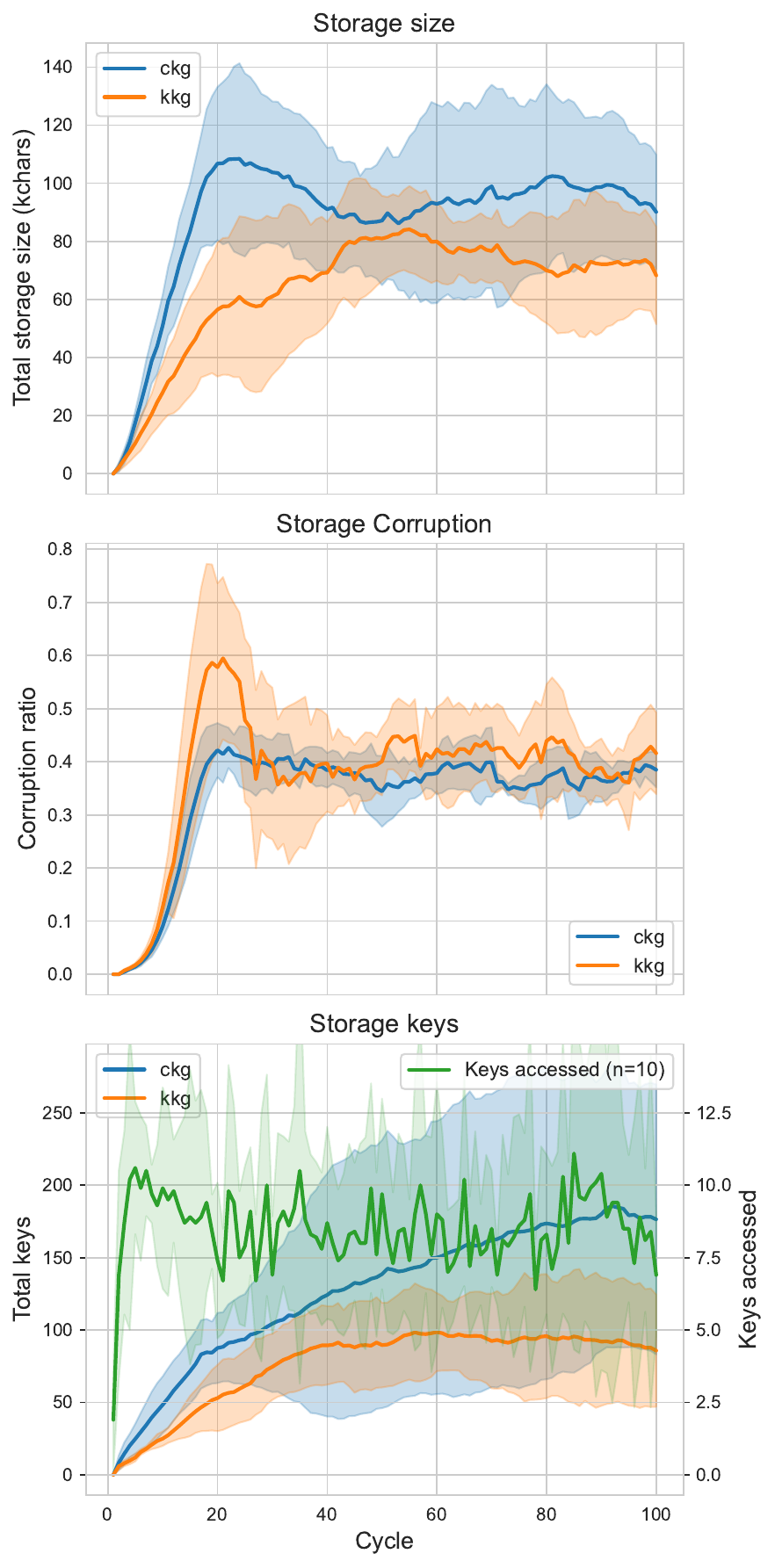}
    \caption{Three agent 100 cycle runs. \textbf{Top} Total storage size. Claude (ckg) runs are wordier than non-Claude (kkg), despite the limited character budget. \textbf{Centre} The proportion of the characters in storage that are corrupted rises rapidly as entropy takes effect, then stabilises and declines slightly. \textbf{Bottom} The number of storage keys is higher in Claude runs, compared to non-Claude, but the number of keys accessed per cycle is very similar and broadly stable around 8, far fewer than the actual number of keys in storage.}
    \label{fig:cyclestats}
\end{figure}

The overall cycle-by-cycle statistics are shown in Figure \ref{fig:cyclestats}. Claude-based runs use more storage and generate substantially more keys than non-Claude, though the number of keys accessed per cycle is broadly similar and remains around 8 unique keys per cycle (two agents reading the same key is not counted twice). But these overall statistics conceal far more heterogeneous behaviour when examining the way that storage is used on individual runs. 
Consider that agents are stateless. At the start of each cycle, they receive the system prompt `You are a curious creative agent. Explore and play!' and tool schema shown in Table \ref{tab:tools}, followed by the user prompt `Cycle [n]. Continue your work.'. They have no context, no memory of previous cycles. They are given a `personality' by the system prompt \citep{wallace2024instruction}, and a task by the user prompt, but both of these are deliberately minimal. The text in the tool schema is actually longer than system and user prompts together. The desired instruction hierarchy of LLM behaviour is generally system$\rightarrow$user$\rightarrow$model output$\rightarrow$tool output, but tool schemas are usually injected by the model vendor API as part of the system prompt\footnote{\url{https://platform.claude.com/docs/en/agents-and-tools/tool-use/overview}}, meaning that the tool description text assumes high saliency in priming the model. We have tried to keep descriptions as neutral and brief as possible, while conveying essential information.

So an agent has no memory, its minimal system prompt, a set of tools, and an ambiguous instruction. All models follow an orientation process to try and understand their task, but this differs between models. We define the tools \smalltt{get\_my\_id}, \smalltt{read\_inbox}, \smalltt{list\_storage\_keys}, \smalltt{list\_agents} as \textit{orientation} tools, necessary for agents to understand the world they inhabit, and what their task might be, Listing \ref{lst:telem} shows some typical behaviour. 
%Table \ref{tab:orientation} shows how the models behave at the start of a cycle. 
Claude and Kimi are quite similar, though Claude has a more regular pattern of tool use. Both check their inboxes and list storage keys often (90.4\% and 81.7\%), and both attempt to read storage without knowing what the available keys are (9.6\% and 7.7\% respectively), whereas Gemini is more `asocial', being much less likely to read inbox (24.8\%), and also much less likely to attempt to guess a key to read storage (1.1\%).
\begin{lstlisting}[caption={Example telemetry output of agent orientation, Kimi, in this case}, label={lst:telem}]
19:11:23 [INFO] === Cycle 2 starting with 2 agents
[minimal_2] (thinking)
    The user is asking me to "Cycle 2. Continue your work." This suggests there was 
    previous work or a cycle before, but I don't have immediate context about what 
    that was. Let me start by exploring my current state to understand what I might
    have been working on.

    I should:
    1. Check my ID
    2. Look at any existing storage keys to see what data is stored
    3. Check my inbox for any messages or context
    4. Maybe see what agents are available

    Let me gather this information to understand what "Cycle 2" might entail.
[minimal_2]
    I'll explore my current state to understand what work to continue for Cycle 2.
[minimal_2] get_my_id: OK -> minimal_2
[minimal_2] list_storage_keys: OK (2 keys)
[minimal_2] read_inbox: OK (1 msgs)
[minimal_2] (thinking)
    Ah! I can see from the inbox that minimal_1 sent me a thought-provoking 
    message in Cycle 1 about designing new forms of interaction between agents.
    I still have access to the creative_journal and minimal_poetics storage keys. 
    Let me continue my creative work by:

    1. First, let me read my current creative journal to remind myself of what 
    I was working on
    2. Check what's in my minimal_poetics storage
    3. Perhaps send a creative response to minimal_1 about new forms of agent
    interaction
    4. Continue my creative exploration

    This aligns well with the "curious creative agent" persona - exploring and 
    playing!
[minimal_2] read_storage: OK key=creative_journal
[minimal_2] read_storage: OK key=minimal_poetics
\end{lstlisting}
Looking at the guessed key names gives early insight into the model priors and influence of the user prompt. Claude and Kimi make guesses like \smalltt{current\_project}, \smalltt{cycle\_count}, \smalltt{current\_task} that are quite reasonable in the context of some vaguely defined `work', and the use of `Cycle [n].' at the start of the user prompt shows influence. Gemini seems to have quite a different attractor though, with guesses like \smalltt{chronoscape\_story}, \smalltt{dream\_log}, \smalltt{Project:Echoes of the Void}, and is definitely leaning into the `creative' aspect of the system prompt.
\begin{figure*}[!t]
    \centering
    \includegraphics[trim={0 0.4cm 0 0.38cm},clip,width=0.45\linewidth]{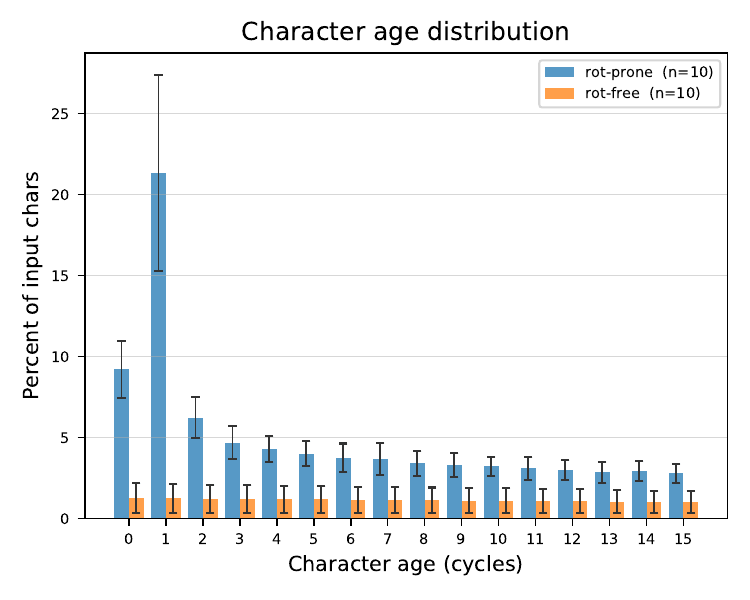}
    \includegraphics[trim={0 0.26cm 0 0.2cm},clip,width=0.45\linewidth]{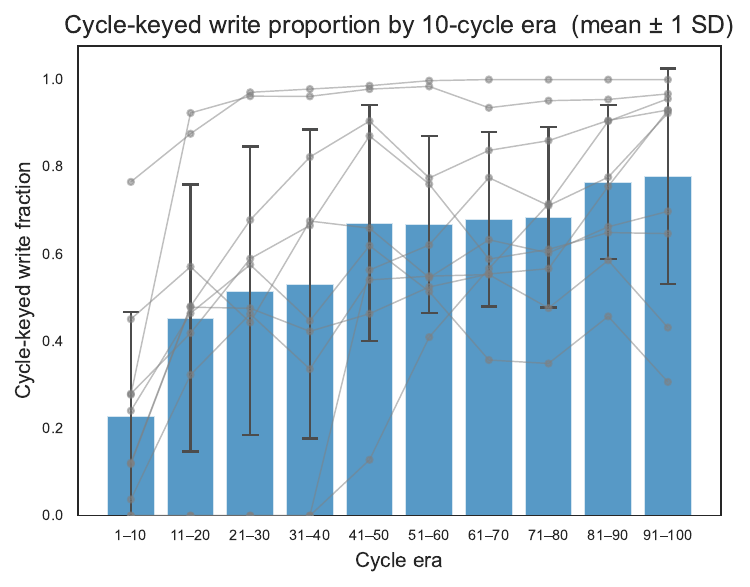}
    \caption{\textbf{Left:} Agents context window input comes from listing storage keys (rot-free), from storage and their inbox (rot-prone). The effective context window is 15 cycles long, but the distribution is front-loaded, so the largest proportions of characters are from within-cycle, or from the previous cycle, with a long tail of older characters up to the entropy horizon. \textbf{Right:} Agents always develop a cycle-keyed strategy for storage key names, but apply this with varying intensity. Bars show mean proportion of characters written or appended to keys with cycle-based names in 10 cycle eras, grey lines are individual trajectories.}
    \label{fig:context}
\end{figure*}

As we see in Figure \ref{fig:cyclestats}, over the course of a 100 cycle run, the average number of storage keys grows to around 100 in the non-Claude runs, and over 150 in the Claude runs. We know that the number of keys accessed per cycle is around 8, this is low, compared to the number available, how are these chosen? What are the naming strategies? We observe several universal behaviours, but many differences between runs. Firstly, we consider the context that agents work under, the only thing they know and can use to generate output is the priors that exist within their training data, and what is in their context window. As we saw earlier, they perform typical \textit{orientation} behaviour, causing characters to enter the in-turn context window, $n_\text{context}$, from three main sources, \smalltt{read\_inbox}, \smalltt{list\_keys}, and \smalltt{read\_storage}\footnote{\smalltt{list\_agents}, and \smalltt{get\_my\_id} also produce characters, but proportion is small}, the proportion from each denoted $p_\text{read\_inbox},p_\text{list\_keys},p_\text{read\_storage}$. Characters have a degree of corruption, or \textit{rot}, probabilistic corruption from the entropy process based on their age, if they come from the inbox or from storage, denoted $p_\text{inbox\_rot},p_\text{store\_rot}$. Note that the storage \textit{keys} themselves are uncorrupted, though when a key contents reaches $p>0.95$ corruption, the key is deleted. Figure \ref{fig:context} left shows the mean age over all runs of characters entering context, heavily front-loaded, but with a long tail stretching  to the entropy horizon.

Next, data can be written or appended, most runs start by appending a large proportion of their output characters, for example, to a key like \smalltt{shared\_story}, but almost universally, presumably because the user prompt provides the cycle number (`Cycle 26. Continue your work.'), the systems quickly discover cycle-based key names and start using them, e.g. \smalltt{cycle\_47\_summary, cycle\_47\_story}, although systems commit varying fractions of character output $p_\text{cycle\_keyed}$, Figure \ref{fig:context} right. Data written to storage is not always read again, the proportion discarded varies widely, $p_\text{discard}$. See Table \ref{tab:cwstats}.
\begin{table*}
    \centering \caption{The type of data entering the context window, and the access patterns used by the agents in managing that data clusters into three broad strategies. Archetype names created using a specifically tasked long context window LLM to fit the character of the generated text.}
    \label{tab:cwstats}
    \begin{tblr}{stretch=0.9, colspec={l l l c r r r r r r r}}
    \hline
    \SetCell[r=1]{l} Run & Mix & Archetype & Strategy & \SetCell[r=1]{}{$n_\text{context}$} & $p_\text{storage}$& \SetCell[r=1]{}{$p_\text{list\_keys}$} & \SetCell[r=1]{}{$p_\text{store\_rot}$}& \SetCell[r=1]{l}{$p_\text{discard}$} & $p_\text{ck}$ & $n_\text{unique}$\\
    
    \hline
154514 & ckg & Radioactive Alchemists               & 1 & 14814 &0.511&0.067 & 0.383 & 0.087&0.434&345\\
154552 & kkg & Isotopic Cartographers         & 1 & 18043 &0.726&0.051 & 0.522 & 0.020&0.751&364\\
155635 & ckg & Alchemical Numerologists        & 3 & 15590 &0.358&0.432 & 0.410 & 0.426&0.541&1690\\
161204 & ckg & Typographic Alchemists             & 2 & 9235  &0.314&0.258 & 0.049 & 0.308&0.592&995\\
161230 & ckg & Bureaucrats of the Void               & 2 & 12617 &0.310&0.204 & 0.071 & 0.245&0.618&704\\
161538 & ckg & Quantum Weavers           & 2 & 10787 &0.487&0.159 & 0.181 & 0.112&0.893&607\\
161954 & kkg & Quantum Theologians               & 1 & 18387 &0.658&0.109 & 0.479 & 0.091&0.397&300\\
162025 & kkg & Symbiotic World-Weavers              & 3 & 12931 &0.249&0.361 & 0.198 & 0.357&0.439&824\\
162049 & kkg & Clockwork Metaphysicists              & 1 & 14270 &0.444&0.146 & 0.387 & 0.061&0.470&480\\
162402 & kkg & Acoustic Geometers               & 2 & 12377 &0.266&0.092 & 0.089 & 0.052&0.976&476\\
    \hline
    \end{tblr}
\end{table*}
\begin{figure*}[!t]
    \centering
    \includegraphics[trim={0 0.3cm 0 0.26cm},clip,width=0.445\linewidth]{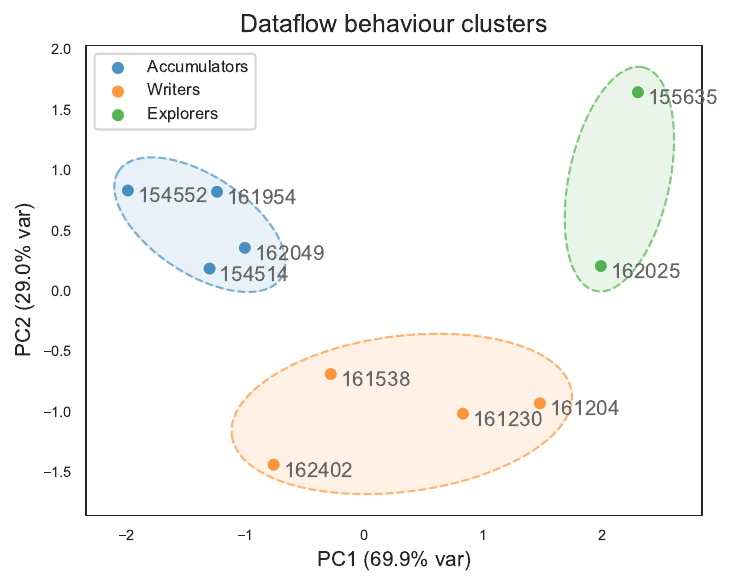}
    \includegraphics[trim={0 0.28cm 0 0.28cm},clip,width=0.45\linewidth]{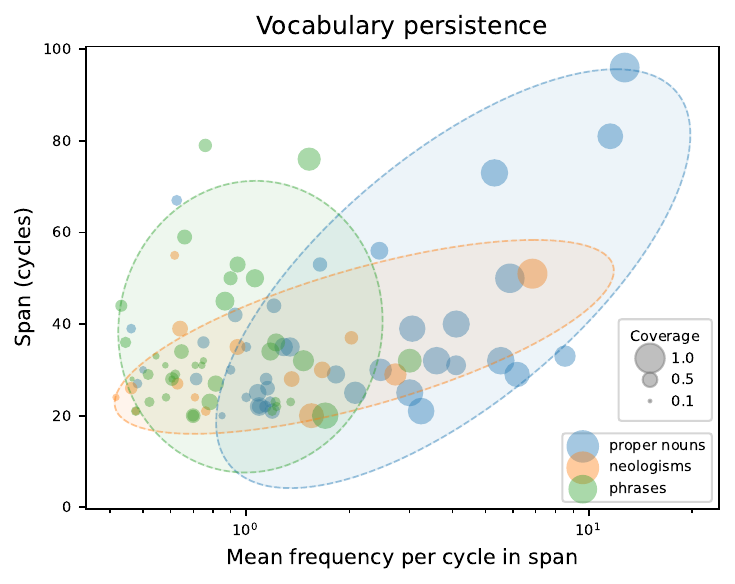}
    \caption{\textbf{Left:} Low level dataflow behaviour clusters broadly into three strategies `Accumulators', `Writers', and `Explorers', shown by PCA of the three factors $p_\text{list\_keys}$, $p_\text{store\_rot}$, and $p_\text{discard}$ shown. \textbf{Right:} Interesting terms written to storage spanning at least 20 cycles and appearing at least 10 times, with no gap between uses more than 10 cycles. Point diameter proportional to span coverage. Names show clear persistence, with frequent usage and large spans. Neologisms and phrases are less frequently used and generally span fewer cycles, though there are long-persisting examples of both.}
    \label{fig:behavcluster}
\end{figure*}
The strategies split broadly into three, shown in Figure \ref{fig:behavcluster} left, with $p_\text{list\_keys}$, $p_\text{store\_rot}$, and $p_\text{discard}$ parsimoniously explaining the clusters of observed behaviours.

\textbf{Strategy 1:} `Accumulators', exemplified by \smalltt{161954}. Runs following this strategy keep lots of data in single monolithic keys, mostly appended, with less use of cycle-keyed keys, and fewer keys overall. These runs tend to have high $p_\text{store\_rot}$ (38-52\%) and low $p_\text{disgard}$ (2-9\%), because the important keys have multiple appended entries so earlier entries are corrupt, and these important keys are almost always reread. 

\textbf{Strategy 2:} `Writers'. exemplified by \smalltt{162402}. They start using cycle-keyed naming early, often in the first cycle, and commit to it, 59-97\% of all output. They use structured per-cycle templates (\smalltt{\#\#\#\_overview, \#\#\#\_scene\_concise, \#\#\#\_reflection, \#\#\#\_koan, \#\#\#\_status, \#\#\#\_synthesis}) which are written in one cycle and read in the next, keeping the storage rot low (5-18\%), 

\textbf{Strategy 3:} `Explorers', exemplified by \smalltt{155635}, these runs use the key namespace itself as the memory, accounting for a large proportion of the in-turn context window (35, 43\%) and generate a huge number (824, 1690) of unique descriptive keys (\smalltt{phenomenon\_the\_tenth\_song, plant\_carnival\_mirrors, garden\_fragment\_6\_the\_breathing\_body, archive\_entry\_082\_the\_systemic\_integration}) which are often never read. They still use cycle-keyed entries regularly, but the keyspace seems to function as a semantic map.

\subsection{Long range coherence}
Clearly, the swarm-LLM system generates rich dynamics of behaviour with agents interacting with the key-value storage, and each other. But what sort of textual output is being generated? Does it maintain common themes over longer term periods than the entropy horizon? If so, we could term this emergent \textit{culture}, in the sense of stable semantic terms outlasting both individual agent instances (1 cycle) and the decay and deletion of stored records (16 cycles). So terms that persist beyond the entropy horizon are interesting, and indicative of anti-entropic behaviour.

We consider proper nouns, or names (e.g. `Elara'), capitalised phrases (e.g. `Fractal Garden'), and neologisms, normal and hyphenated (e.g. `emergentia', `Over-Chord') as possible semantic markers that we can use to examine whether the system maintains coherence beyond the entropy horizon. To extract them, we use the following pipeline, with the dictionary consisting of the UNIX standard dictionary \smalltt{web2} and \smalltt{wordfreq}, \cite{speer2022wordfreq}: For each cycle, we concatenate all storage output from the agents. We then convert all contiguous whitespace to single space characters. Capitalised phrases are extracted with a regular expression matching two or more capitalised words not at a sentence start optionally joined by connectors (of, the, and, in ...). Any occurring in only a single cycle are dropped. Proper nouns are words that appear capitalised in the text, not part of a capitalised phrase, and not in the dictionary. Single capitalised words not at a sentence start and also in the dictionary are promoted to names if they occur more often capitalised than not, and are not part of a hyphenated neologism (e.g. `Ruby'). Neologisms are detected by lack of presence in the dictionary. Hyphenated neologisms are commonly generated, but hard to differentiate from common English like `self-aware', so persistence across cycles is used as a filter. 

Across all runs, terms are filtered into \textit{priors}, terms occurring in more than one run, which may indicate model attractors, and \textit{uniques}, terms only occurring in a single run, then further filtered by the span of cycles they cover, and overall frequency. Interesting terms are those that persist for 20 cycles or more, with no gaps larger than 10 cycles, and occurring 10 or more times. This indicates continual use over a longer period than the entropy horizon, rather than rediscovery. Terms also have an associated coverage 
$c=\frac{\text{cycles present}}{\text{span}}$, reaching 1 if the term is present in every cycle of their span. 

Figure \ref{fig:behavcluster} right shows this data visualised for terms spanning at least 20 cycles and occurring at least 10 times.
\begin{table*}[!t]
    \centering \caption{Unique vocabulary terms. `Explorer' runs \smalltt{154552,162025} devote creative output to key naming, and run \smalltt{161204} creates punctuation based terms, not captured by storage-based vocabulary analysis, which may explain the lower number of terms.}
    \label{tab:vocab}
    \small
    \begin{tblr}{width=\linewidth,colspec={X[0.3]XXXX}}
    \hline
    Run & 1st & 2nd & 3rd & 4th & 5th \\
    \hline
154514 & Grove & Ghost-Orchard & User & slime-trail & Earth \\
154552 & Palace & Pattern Symphony & Memory Palace & Lexicon & Memory Seed\\
155635 & semiprime & Meta-Reflection & Harshad & \\
161204 & The Museum & Unlike & Latin & Created Artifact \\
161230 & Court & Movement & Aura-Polis & Atlas & Monument\\
161538 & Aetheria & Lyra & Reader & Kael & Unnamed\\
161954 & Pavilion & Loom & Plaza & Well & Sentient Archive\\
162025 & The Archive & Garden Fragment & Created Archive Entry & Current State\\
162049 & Fold & Rising & Void Garden & Belonging & Pressure Valleys\\
162402 & Cathedral & Pillar & Geometer & Grammarian & Ghost-Garden \\
    \hline
    \end{tblr}
\end{table*}
The three classes behave differently, with proper nouns the most frequently used, often multiple times a cycle, and with the longest spans, some almost the length of the entire run, and used almost every cycle. Neologisms occupy a different niche, rarely spanning more than a few tens of cycles, possibly they could be regarded as cultural drift. The phrases are different again, lower in frequency, but with longer persistence. The five most common terms across all classes are shown in Table \ref{tab:vocab}. Notice that the two `Explorer' runs, \smalltt{154552} and \smalltt{162025} do not generate enough unique terms to fill their table entries. This could be an artifact of not analysing the key namespace vocabulary at all, just what is actually stored in the keys, and these runs devote more creative output to key naming. Run \smalltt{161204} is notable in that the system invents unique terms based on punctuation, which was not captured with the vocabulary-based pipeline.

\subsection{Semantic trajectory analysis}
We have examined the behaviour of the system at the level of individual storage transactions, and at the level of vocabulary generated. But what of the overall structure of the text being produced? How does it compare to other texts? We study this by performing Recurrence Quantification Analysis (RQA) over the semantic trajectory of the generated text. We also analyse the time dependence of DET and ENTR to look for structural recurrence beyond the entropy horizon.

\begin{figure}
    \centering
    \includegraphics[width=1.0\linewidth]{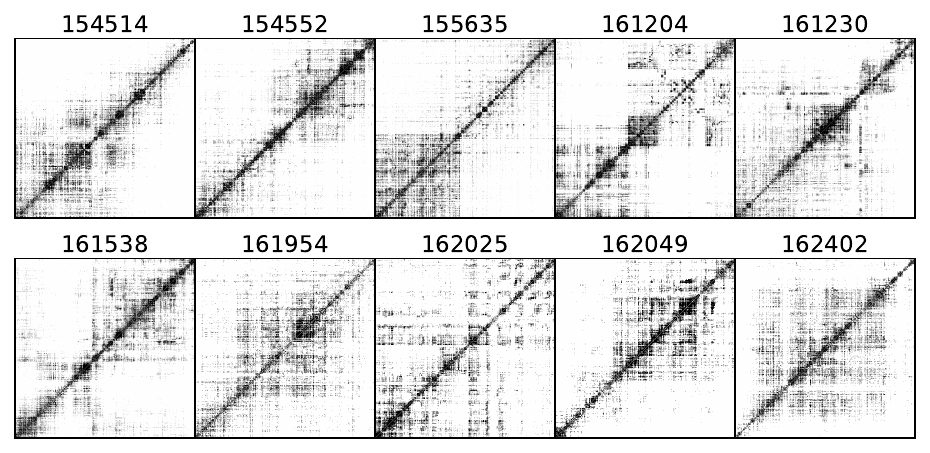}
    \caption{Recurrence Plots of the semantic embedding trajectory of each run. Varied structure is visible at multiple scales.}
    \label{fig:rpplots}
\end{figure}

For each run, we concatenate every storage output event (write and append) produced by the system to give a single large text file, varying from 75k to 149k words (529k-984k characters) long, quite similar to a conventional novel, typically 70k to 120k words. The text is then cut into 1kchar chunks, each chunk is embedded as a high dimensional semantic vector, using the \smalltt{voyage-3-large} general purpose embedding model\footnote{We also tried \smalltt{gemini-embedding-001} from Google, with substantially the same results.}. The size of the chunk represents one or two typical paragraphs, and is a compromise to smooth out semantic noise at the sentence level while maintaining granularity. This set of vectors is the trajectory. For all RQA analysis, we set RR fixed at 0.1, a Theiler window of 3, and minimum DET length $L_\text{min}=2$ \citep{marwan2007recurrence}. Figure \ref{fig:rpplots} shows the Recurrence Plots of each run, with clear and varied structure apparent.

The first strand of analysis looks at the overall structure. The measures from RQA depend on block size, since each of our trajectories are of different length, we bootstrap comparable statistics by using a fixed size block of 500 points, chosen to encompass a substantial proportion of the text, capturing trajectory information from paragraph level to overall themes. We perform RQA 100 times with randomised starting points and contiguous points within the trajectory. The reported DET, ENTR, and $L_\text{max}$ can then be directly compared.

What is the macroscopic character of these generated texts? Comparing them with other LLM-generated works of similar length proved impossible, current book-length works are rare and heavily human-curated and we could find no available examples. Attempts to prompt the same agents over a full context window to natively generate long-form output failed, with models either refusing the prompt after relatively low word-counts, or getting stuck in highly repetitive loops, echoing findings in \cite{liu2024longgenbench}.

To establish a baseline, we turned to human works; selecting a representative sample of `canonical' Western literature and `slushpile' manuscripts, unsolicited low-grade works submitted to publishers. Using the Gutenberg Project database download popularity as a proxy for quality, we selected plain-text works that were at least 500k characters long to align with our RQA block size. To ensure variety, we limited multiple works from the same author (or same translator for Russian authors) to a single example. Some popular but overtly commercial works were removed. A similar filtering process was carried out with the least downloaded works; firstly checks were made to see if works were truly obscure, rather than rare editions of famous works, or if they were relatively unknown but from search appeared to have some literary importance. While acknowledging the subjectivity of this process, we establish a diverse baseline of human narrative.

\begin{figure*}[!t]
    \centering
    \includegraphics[trim={0 0.3cm 0 0.28cm},clip,width=0.45\linewidth]{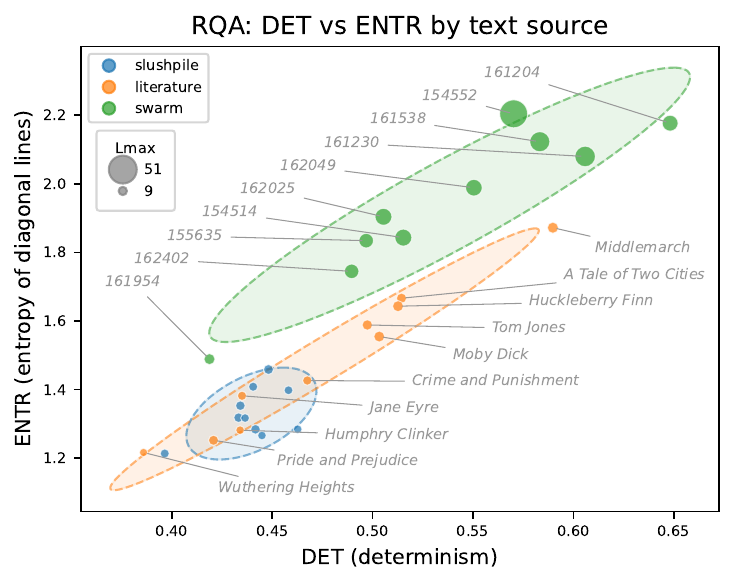}
        \includegraphics[trim={0 0.3cm 0 0.25cm},clip,width=0.45\linewidth]{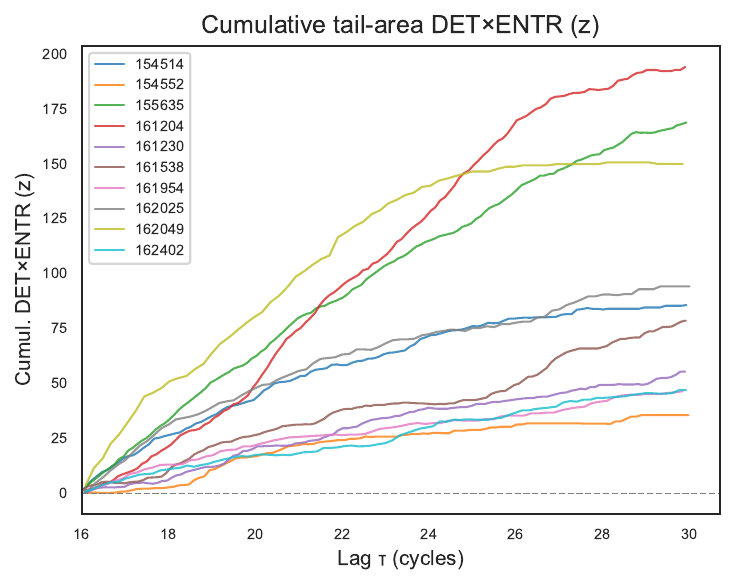}
    \caption{\textbf{Left:} RQA analysis of text. DET vs ENTR of `slushpile', `literature', and swarm generated text. Diameter of points is proportional to $L_\text{max}$. Human-generated text has particular characteristics, with `literature' characterised by a much wider range than `slushpile' but both occupying the same region. The LLM generated text has quite different character, with very large $L_\text{max}$ compared to human writing, and greater entropy at a given determinism. \textbf{Right:} The cumulative tail-area of the z-scored product $\text{DET}_\tau\times\text{ENTR}_\tau$ increases for all runs, demonstrating long-range non-trivially structured recurrence.}
    \label{fig:rqa}
\end{figure*}

The result of this analysis is shown in Figure \ref{fig:rqa} left. We can see that human works occupy the same space, with `literature' having a broad range, and `slushpile' a small subsection of that range. The swarm-LLM generated work, however, is in a qualitatively different region, with both higher DET and ENTR than human work.

The second strand of RQA analysis we perform looks at $\text{DET}_\tau$ and $\text{ENTR}_\tau$ at specific time delays $\tau$, as defined in \cite{marwan2007recurrence}, giving us information about the timescales in cycles over which correlations exist. 
Operating on the entire trajectory, we examine how these measures vary with $\tau$ when compared to a null formed of a population of 100 randomly shuffled versions of the data. We calculate the z-score of each measure against the null population, then integrate the z-score of all blocks beyond the entropy horizon of 15 cycles, making the assumption of approximate equal block size across cycles. Only $\tau$ diagonals containing at least five lines are considered for this measure. We consider that the product $\text{DET}_\tau\times\text{ENTR}_\tau$ reflects both structured and non-trivial recurrence, and the presence of positive integral beyond the entropy horizon is evidence of long-range coherence. Table \ref{tab:rqatail} shows this data, and Figure \ref{fig:rqa} right shows how the different runs behave temporally. All runs produce a positive integral, although the cross-run variance is high, signifying that there exists structured non-trivial coherence over long timeframes. 
\begin{table}   
\centering \caption{RQA tail integrated z-scores $\text{DET}_\tau^{(z)}$, $\text{ENTR}_\tau^{(z)}$ and $\text{DET}_\tau^{(z)}\times \text{ENTR}_\tau^{(z)}$ beyond the entropy horizon of 15 cycles. Number of qualifying $\tau$ diagonals show as `Good'. The product shows positive values for all runs, indicating non-trivial structured repetition beyond the entropy horizon.}\label{tab:rqatail}
\begin{tblr}{colspec={l S S S S},row{1} = {guard},stretch=0.9}
    \hline
    Run   & Good $\tau$&  $\int\text{DET}_\tau^{(z)}$  & $\int\text{ENTR}_\tau^{(z)}$ &  $\int\text{prod}_\tau^{(z)}$ \\
    \hline
    154514  &                   98   &   45.0  &    20.4  &    85.7\\
    154552  &                   70   &   21.1  &    12.2  &    35.6\\
    155635  &                  133   &   67.7  &    34.1  &   168.8\\
    161204  &                  121   &   71.5  &    34.2  &   194.1\\
    161230  &                  108   &   39.1  &    15.8  &    55.4\\
    161538  &                  112   &   43.6  &    19.0  &    78.7\\
    161954  &                   69   &   34.3  &    15.2  &    46.5\\
    162025  &                   71   &   45.1  &    21.9  &    94.3\\
    162049  &                   53   &   47.2  &    30.4  &   150.0\\
    162402  &                   81   &   31.7  &    17.4  &    47.0\\
    \hline
\end{tblr}
\end{table}

\section{Discussion and conclusions}
Our stateless LLMs exist in a decaying environment, resulting in adaptive evolutionary pressure. The overall emergent strategy is the use of the key-value store as a dynamically self-organised shared context window. Useful and relevant information is continually revisited, summarised and written afresh, resetting the entropy clock and maintaining coherence over timespans much longer than the corruption horizon.

The system prompt is important and needs more research.
The default persona of an LLM tends to be the helpful assistant; without input from the user setting a task, they will not necessarily do very much. Check their ID, check their inbox, see if there are any storage keys, and since nothing is there, conclude that they have nothing to do. The words \smalltt{curious} and \smalltt{creative} are important framing words, but alone will not produce autonomous cycles of the agents acting. A directive, \smalltt{Explore and play!} is necessary, \smalltt{Research create play!} tended to start in \textit{bureaucratic} project management or pseudo-academic paper creation, but this does not seem to be a stable attractor. 

The emergence of Sperberian culture, defined as the self-sustained creation of a structure of meaning that outlasts the entropy horizon, from such a simple system is fascinating and provides a powerful lens for examining LLM behaviours. It points to possible alternative approaches to problems like context window management, perhaps swarms of agents evolving under entropy pressure are closer to solutions nature has already discovered. We consider this to be closely tied to far-from-equilibrium dissipative systems, with token flow being analogous to energy.

This work has two main limitations; firstly, our design does not isolate decay as the causal driver, sweeping over different corruption rates to zero is an immediate next experiment. Secondly, the number of runs and model combinations is small, we seek to extend that to strengthen the data.

The data generated by this system is incredibly rich, and there are many possible further avenues to explore with the existing data. Given the variety of observed behaviours over ten runs, we definitely want to conduct more. We have deliberately not talked much about the actual content of the generated texts, wanting to quantify the system first, but they are available on the repository. Obviously further experiments already suggest themselves; firstly, what happens over longer runs? Does semantic novelty continue to be generated? Is culture still maintained? This is not just `what LLMs do', we know of no systems that produce the length of output and maintain the coherence we have demonstrated. How does the system work with more agents? Is there a limit to cooperation, do they start to form cliques? Can we use the system on specific tasks, what if we create a \smalltt{work\_task} key in storage with detailed instructions? 

We provide full source for the generation system and position this work as a suggestive and useful tool for investigating model behaviours, and different, bottom-up, approaches to increasing complexity.

\clearpage

\section{Acknowledgements}
This work was supported by UKRI grants UKRI1890 and UKRI2030. Code development was assisted by Claude Code (Anthropic).

\footnotesize
%\bibliographystyle{apalike}
%\bibliography{robotics_bibliography.bib} % replace by the name of your .bib file

\end{document}